%% file: acl_latex.tex
\documentclass[11pt]{article}

\usepackage[preprint]{acl}

\usepackage{times}
\usepackage{latexsym}

\usepackage[T1]{fontenc}

\usepackage[utf8]{inputenc}

\usepackage{microtype}

\usepackage{inconsolata}

\usepackage{graphicx}

\usepackage{amsmath}
\usepackage{amssymb}
\usepackage{booktabs}
\usepackage{multirow}
\usepackage{tabularx}
\usepackage[table]{xcolor}
\title{Accelerate Speculative Decoding with Sparse Computation in Verification}

\author{
 \textbf{Jikai Wang\textsuperscript{1,}\textsuperscript{2}},
 \textbf{Jianchao Tan\textsuperscript{2$\dagger$}},
 \textbf{Yuxuan Hu \textsuperscript{2}},
 \textbf{Jiayu Qin\textsuperscript{2}},
\\
 \textbf{Yerui Sun\textsuperscript{2}},
 \textbf{Yuchen Xie\textsuperscript{2}},
 \textbf{Xunliang Cai\textsuperscript{2}},
 \textbf{Juntao Li\textsuperscript{1$\dagger$}},
 \textbf{Min Zhang\textsuperscript{1}},
\\
 \textsuperscript{1}Key Laboratory of Data Intelligence and Advanced Computing, Soochow University\\
 \textsuperscript{2}Meituan\\
    \href{mailto:risus254@gmail.com}{risus254@gmail.com},
    \href{mailto:tanjianchao02@gmail.com}{tanjianchao02@meituan.com},
    \href{mailto:ljt@suda.edu.cn}{ljt@suda.edu.cn}
 }

\begin{document}
\maketitle
\begin{abstract}
Speculative decoding accelerates autoregressive language model inference by verifying multiple draft tokens in parallel. However, the verification stage often becomes the dominant computational bottleneck, especially for long-context inputs and mixture-of-experts (MoE) models.
Existing sparsification methods are designed primarily for standard token-by-token autoregressive decoding to remove substantial computational redundancy in LLMs.
This work systematically adopts different sparse methods on the verification stage of the speculative decoding and identifies structured redundancy across multiple dimensions.
Based on these observations, we propose a sparse verification framework that jointly sparsifies attention, FFN, and MoE components during the verification stage to reduce the dominant computation cost.
The framework further incorporates an inter-draft token and inter-layer retrieval reuse strategy to further reduce redundant computation without introducing additional training.
Extensive experiments across summarization, question answering, and mathematical reasoning datasets demonstrate that the proposed methods achieve favorable efficiency-accuracy trade-offs, while maintaining stable acceptance length.
\end{abstract}

\footnotetext{Corresponding authors.}

\section{Introduction}
Autoregressive large language models (LLMs) \citep{llama31,openai2024gpt4technicalreport,guo2025deepseek} have achieved remarkable success across a wide range of tasks, but their inference cost continues to grow rapidly with increasing model size and context length. While training-time optimizations have received extensive attention, inference efficiency remains a critical bottleneck, especially in latency-sensitive and long-context scenarios.

\input{tables/flops}

Speculative decoding \citep{stern2018blockwise,leviathan2023fast,xia2023speculative} has recently emerged as an effective technique to accelerate autoregressive generation without altering the output distribution. By introducing a lightweight draft model to propose multiple candidate tokens and verifying them in parallel using the target model, speculative decoding amortizes decoding overhead across multiple tokens per step. However, this acceleration shifts the computational burden to the verification stage. In particular, when draft lengths grow, or long-context inputs are involved, verification incurs substantial overhead due to full attention over large KV caches, dense feed-forward computation, and multi-expert evaluation in MoE models. As a result, verification becomes the dominant bottleneck in speculative decoding.

\begin{figure*}
    \centering
    \includegraphics[width=\textwidth]{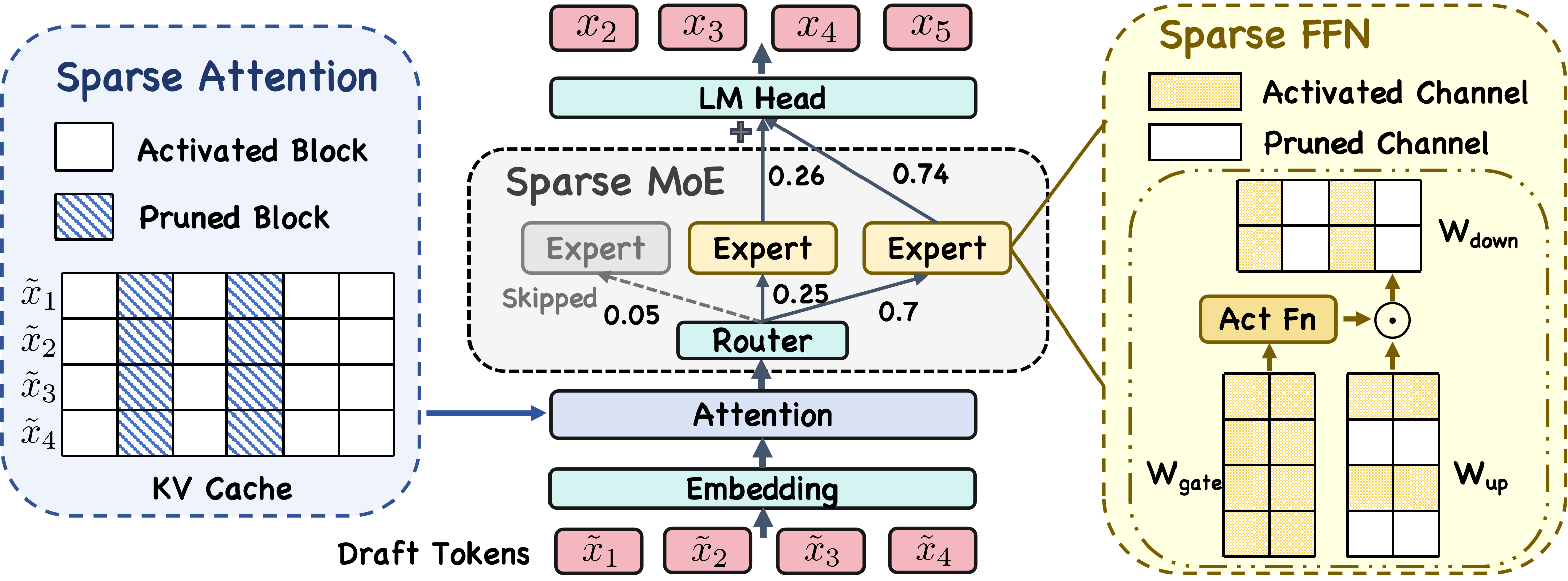}
    \caption{An illustration of the three sparse verification strategies within a single Transformer layer. Residual connections and normalization layers are omitted for clarity. A KV block contains the key-value states of multiple prefix tokens. In sparse attention, the selection of blocks is determined by the dot product between the query of the first draft token and a representative value of each block. Multiple draft tokens select the same set of blocks. In sparse FFN, the pruned channels are determined by the output of the activation function and a given threshold. In sparse MoE, a dynamic strategy based on the weights of the selected experts determines whether an expert participates in computation, and different draft tokens may involve different numbers of experts. }
    \label{fig:intro}
\end{figure*}

Recent advances in sparse inference have demonstrated that significant redundancy exists in different components of LLMs. Sparse attention and KV cache eviction techniques reduce memory and computation by selectively retaining important context tokens. Sparse FFN methods exploit activation sparsity to reduce matrix multiplications. MoE models further reduce computation by activating only a small subset of experts per token. However, existing sparse inference methods are primarily designed for standard autoregressive decoding, acknowledging only a single token per step. Their application and effects on speculative decoding are not explored.

In this work, we systematically study sparse verification for speculative decoding. We observe that verification exhibits structured redundancy across multiple dimensions, including attention computation, feed-forward activation, expert utilization, and even across transformer layers. Building on these observations, we propose a unified sparse verification framework that applies sparsification jointly to attention, feed-forward networks, and MoE layers, while respecting the unique requirements of speculative decoding.

Specifically, we introduce (1) an importance-based sparse attention mechanism tailored for multi-token verification, combined with a piecewise budget control strategy to ensure stability on short contexts; (2) an inter-layer retrieval reuse scheme that avoids redundant block selection across similar layers; (3) a sparse feed-forward verification method that prunes low-activation channels during inference; and (4) a generalized sparse MoE strategy that adaptively skips low-contribution experts. We further combine these techniques into a hybrid sparse verification method that sparsifies verification along three orthogonal dimensions.
We provide a detailed FLOPs analysis of the proposed methods in Table \ref{tab:flops}, showing that sparse verification substantially reduces the dominant computation terms in attention, FFN, and MoE layers. Extensive experiments across long-context, question answering, and mathematical reasoning benchmarks demonstrate that our methods achieve significant efficiency gains while maintaining competitive verification accuracy and stable acceptance length.

\section{Preliminary}
\subsection{Speculative Decoding}
Speculative decoding \citep{stern2018blockwise,leviathan2023fast,xia2023speculative} is an inference technique for autoregressive models that aims to accelerate generation.
Let $p_\theta(x_t \mid x_{<t})$ denote the conditional distribution of the target token $x_t$ given previous tokens $x_{<t}$ under the target model with parameters $\theta$.  
Speculative decoding introduces a lightweight draft model, with parameters $\phi$, to propose multiple candidate tokens in advance.
At time step $t$, the draft model generates $K$ speculative tokens $(\tilde{x}_{t+1}, \tilde{x}_{t+2}, \dots, \tilde{x}_{t+K})$ sequentially:
\begin{equation}
\tilde{x}_{t+k} \sim p_\phi(x_{t+k} \mid x_{<t+k}), \quad k = 1, \dots, K.
\end{equation}

In the verification phase, the target model computes the probabilities 
$P_{t:t+K} = p_\theta(\tilde{x}_{t:t+K} \mid x_{<t})$ 
for the draft tokens $\tilde{x}_{t:t+K}$.  
Each draft token $\tilde{x}_{t+k}$ is then sequentially evaluated with an acceptance probability:
\begin{equation}
\alpha_{t+k} = \min \Big( 1, \frac{p_\theta(\tilde{x}_{t+k} \mid x_{<t+k})}{p_{\phi}(\tilde{x}_{t+k} \mid x_{<t+k})} \Big),
\end{equation}
where $p_{\phi}$ is the draft model's output probability.  
Note that once a token $\tilde{x}_{t+k}$ is rejected, all its subsequent draft tokens are discarded.
The rejected token is then resampled from the adjusted distribution:
\begin{equation}
\hat{p} =norm( \max\big(0, p_\theta - p_{\phi}\big)),
\end{equation}
ensuring that the overall generation remains consistent with the target model distribution.

\subsection{KV Cache Eviction in Sparse Verification}
\begin{figure}
    \centering
    \includegraphics[width=\linewidth]{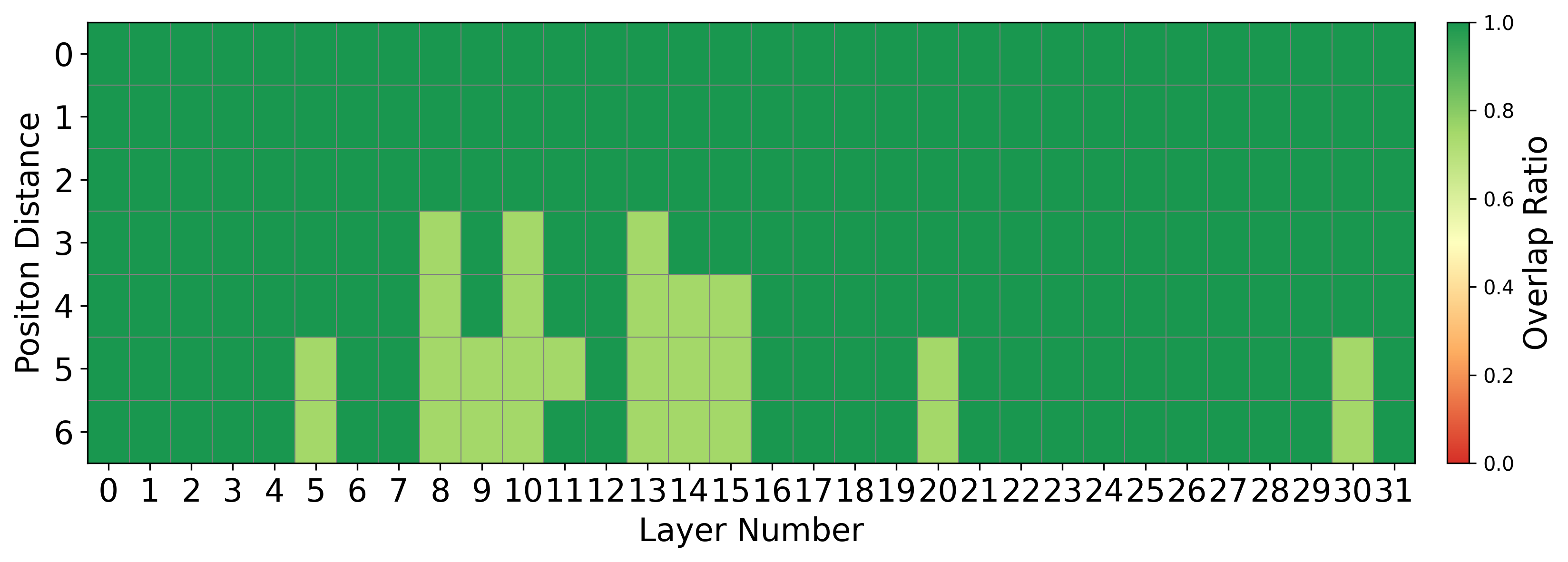}
    \caption{The overlap ratio of the KV block selection for the draft token pairs with different position distances.}
    \label{fig:block_overlap}
\end{figure}

A major trend in key-value cache eviction is to determine which caches to activate based on the current token information \citep{tang2024quest,lu2025mobamixtureblockattention}.
For speculative decoding, each step of the decoding process involves multiple tokens as input. Therefore, it is necessary to analyze the retrieval differences of these tokens to better adapt to these strategies.
To this end, we perform speculative decoding on Llama3.1-8B \citep{grattafiori2024llama3herdmodels} and analyze the overlap of cache blocks retrieved by different draft tokens based on their respective query values.
Figure \ref{fig:block_overlap} displays the results.
In the experiments, a tree-structured draft is adopted, with 60 draft tokens in total. We measure the overlap ratio of retrieved blocks between token pairs at different positional distances. The overlap is calculated as the intersection of the retrieved blocks for the two tokens divided by the total number of blocks.
A distance of 0 indicates two different tokens located at the same position.

The results show that tokens within the same step retrieve highly similar blocks, with an average overlap exceeding 0.8 in most layers. Moreover, tokens that are closer in position tend to retrieve more similar blocks. On the other hand, in the draft tree, the first token is the accepted token sampled from the previous step, making it more important than the others since it influences the verification of all subsequent tokens. Taking into account both aspects, we use the first token to perform block retrieval, while the other tokens reuse its retrieved blocks during verification.

\section{Sparse Verification}
In this section, we introduce a sparse verification framework that exploits redundancy in the verification process. The key idea is that verification does not require full computation across all model components. We therefore apply controlled sparsification along three dimensions: attention, feed-forward networks, and mixture-of-experts layers.
Unlike prior sparsity-aware methods \citep{kwon2022fastposttrainingpruningframework,yuan2025native} that require retraining or architectural modification, our approach operates entirely at inference time. This design choice allows sparse verification to be directly applied to off-the-shelf large models.

\subsection{Sparse Attention}

Existing sparse attention methods primarily focus on the decoding stage by applying KV cache eviction to reduce memory and computation overhead. However, such strategies are designed for standard autoregressive decoding, where each step processes only a single token.
In speculative decoding, by contrast, each step must verify multiple candidate tokens simultaneously, often involving several to dozens of tokens. This multi-token verification introduces new challenges for efficiently selecting relevant context while maintaining verification accuracy.

To address this, we introduce a lightweight importance-based sparse attention mechanism tailored for speculative decoding. Inspired by recent block-wise retrieval ideas as Quest \citep{tang2024quest}, we partition the current KV cache into structured blocks according to both token positions and KV heads.

Formally, each block contains all the KV entries corresponding to one attention head and a fixed number of consecutive tokens, denoted as the block size $B$.
If there are $H$ KV heads and a total sequence length of $L_{\text{seq}}$, the KV cache can be represented as:
\begin{align}
\mathcal{K} &= \{\mathbf{K}_{h,b}\}, \quad 
\mathcal{V} = \{\mathbf{V}_{h,b}\}, \\
h &\in [1,H], \; b \in [1, \lceil L_{\text{seq}} / B \rceil]. \nonumber
\end{align}

For speculative verification, we compute an importance score for each block using the query of the first draft token $\mathbf{q}_0$:
\begin{equation}
s_{h,b} = \frac{1}{B} \sum_{\mathbf{k} \in \mathbf{K}_{h,b}} \mathbf{q}_0^\top \mathbf{k},
\end{equation}
and select the top-$N$ blocks with the highest scores for verification.

Moreover, due to the existence of the attention sink \citep{xiaoefficient}, where early tokens tend to attract disproportionate attention across layers, and the locality bias \citep{yang2018modeling,wuemergence}, where neighboring tokens play a more critical role in contextual reasoning, we retain both the first and last few blocks without eviction.
This boundary-preserving design prevents the loss of essential contextual information that is frequently referenced during verification.

\subsection{Piecewise Budget Control}

In practice, we observe that when the KV cache length is below a certain threshold, the verification time does not grow noticeably with the cache size.
Moreover, for short-text scenarios, aggressive KV eviction may introduce unnecessary randomness and performance degradation.
To balance efficiency and stability, we adopt a piecewise budget control strategy.

Specifically, when the sequence length $L_{\text{seq}}$ is below a fixed threshold $L_0$, we perform no eviction.
When $L_{\text{seq}} > L_0$, the number of retained blocks is adaptively determined by:
\begin{equation}
N_{\text{budget}} = \frac{\left((L_{\text{seq}} - L_0) \times \rho + L_0\right) \times H}{B},
\end{equation}
where $0 < \rho < 1$ is a sparsity coefficient controlling the proportion of active blocks,
$H$ is the number of KV heads, and $B$ is the block size.

This design ensures that for short sequences, verification remains stable and deterministic, while for longer contexts, computational cost scales smoothly with sequence length under a controlled sparsity budget.

\subsection{Inter-layer Retrieval Reuse}
\input{tables/sparse_attn}

Through empirical analysis of the retrieved blocks across transformer layers, we observe that although the retrieval patterns in shallow layers tend to be irregular and diverse, the block selection in middle and deeper layers exhibits strong inter-layer similarity, particularly among adjacent layers.
This observation suggests that performing retrieval at all layers is redundant and leads to unnecessary computational overhead.

To exploit this redundancy, we introduce an inter-layer retrieval reuse strategy that selectively performs retrieval only on representative layers.
We first construct a set of calibration data $\mathcal{D}_{\text{cal}}$ covering various input lengths to analyze the model's retrieval behavior.
For each input sample, we record the mask of the blocks retrieved for each layer $l$, denoted as $M_l \in \{0,1\}^{N_b}$, where $N_b$ is the total number of candidate blocks.

We then compute the pairwise retrieval similarity between adjacent layers using the Jaccard similarity:
\begin{equation}
J(M_l, M_{l-1}) = \frac{|M_l \cap M_{l-1}|}{|M_l \cup M_{l-1}|}.
\end{equation}
For the first layer, we set $J(M_1, M_0) = 0$ by definition.

Based on these similarity scores, we identify a subset of layers with the lowest similarity to their preceding layers as anchor layers:
\begin{equation}
\mathcal{A} = \text{TopK}_{l}(1 - J(M_l, M_{l-1})),
\end{equation}
where $|\mathcal{A}| = K$ and $K$ is the number of anchor layers to be selected.

Only the anchor layers perform block retrieval to determine which blocks participate in attention computation.  
For non-anchor layers, we directly reuse the retrieval results of the nearest preceding anchor layer:
\begin{equation}
M_l = M_{a(l)}, \quad a(l) = \max\{i \in \mathcal{A} \mid i < l\}.
\end{equation}

This layer-wise reuse mechanism effectively reduces retrieval operations while maintaining comparable verification quality, as the selected anchor layers capture the essential variation of retrieval patterns across the network.

\subsection{Sparse Feedforward Network}

Existing sparse FFN (SFFN) methods \citep{roller2021hash,liu2023towards} are primarily designed for the training stage, where sparsity is introduced to reduce the cost of updating parameters.  
Inspired by these approaches, we leverage the inherent sparsity of the feed-forward network (FFN) during inference to reduce the computational cost of speculative verification.

A standard gated FFN consists of an gate-projection, an up-projection, an activation function, and a down-projection.
Let the input to the FFN at layer $l$ be $\mathbf{x}_l \in \mathbb{R}^d$.  
The FFN computation can be written as:
\begin{equation}
\mathbf{h}_l = \sigma(W_{\text{gate}} \mathbf{x}_l + \mathbf{b}_{\text{gate}}), 
\end{equation}
\begin{equation}
\mathbf{y}_l = W_{\text{down}} (\mathbf{h}_l \odot (W_{\text{up}} \mathbf{x}_l+\mathbf{b}_{\text{up}}))+ \mathbf{b}_{\text{down}},
\end{equation}
where $\sigma(\cdot)$ denotes the activation function, typically GeLU \citep{hendrycks2016gaussian} or SwiGLU \citep{shazeer2020glu}.

We observe that many channels in the activated hidden vector $\mathbf{h}_l$ contribute negligibly to the final output.  
To exploit this sparsity, we identify inactive channels by thresholding the activation magnitude:
\begin{equation}
\mathcal{S}_l = \{ i \mid |h_{l,i}| < \tau \},
\end{equation}
where $\tau$ is a predefined threshold and $h_{l,i}$ is the $i$-th dimension of $\mathbf{h}_l$.

Only channels not belonging to $\mathcal{S}_l$ are used in the up and down projection.  
Thus, the sparse FFN output becomes:
\begin{equation}
\mathbf{h}_{\text{up}} = \mathbf{h}_l^{(\neg \mathcal{S}_l)} \odot (W_{\text{up}}^{(\neg \mathcal{S}_l)} \mathbf{x}_l^{(\neg \mathcal{S}_l)}+\mathbf{b}_{\text{up}}^{(\neg \mathcal{S}_l)}),
\end{equation}
\begin{equation}
\mathbf{y}_l = W_{\text{down}}^{(\neg \mathcal{S}_l)} \mathbf{h}_{\text{up}}+ \mathbf{b}_{\text{down}}^{(\neg \mathcal{S}_l)},
\end{equation}
where $(\neg \mathcal{S}_l)$ indicates the set of active channels.  
This mechanism preserves the structure of the FFN while significantly reducing multiplications in the up-projection and down-projection, which dominates the FFN computational cost.
By applying this sparsification during speculative verification, we reduce overall inference cost while maintaining comparable validation accuracy.

\subsection{Sparse Mixture of Experts}
We also propose an adaptive expert skipping method for each candidate token for MoE target models.
\citet{lu-etal-2024-experts} have introduced a dynamic skipping expert strategy during inference.
They consider the case of $k{=}2$ activated experts, where the second expert may be skipped based on a threshold determined by the ratio of routing logits. We generalize this mechanism to arbitrarily activated expert counts $k>2$, and allow skipping up to $m$ experts ($1 \le m < k$) for each token during inference.

Given $k$ routed experts in an MoE layer, we denote their routing weights, after sorting in ascending order, as $w = \{w_{1}, w_{2}, \ldots, w_{k}\}$.
To determine how many experts can be skipped, we compute a layer-specific threshold for every possible skip size $m$. For a calibration dataset, we first calculate a ratio for all calibration tokens:
\begin{equation}
\frac{\sum_{j=1}^{k-m} w_{j}}{\sum_{j=1}^{k} w_{j}},
\end{equation}
and define the threshold $\beta_{m}$ as the median of these ratio values in that layer. This produces a \emph{threshold map} $\{\beta_{1},\beta_{2},\ldots,\beta_{k-1}\}$ that specifies allowed skip levels for that layer.

During inference, for a given token, we search for the maximum number of experts $i$ to skip ($0 \le i \le m$) such that
\begin{equation}
    \frac{\sum_{j=1}^{k-i} w_{j}}{\sum_{j=1}^{k} w_{j}} < \beta_{m}.
\end{equation}
If such $i>0$ exists, we skip the $i$ lowest-weight experts $\{e_{1}, e_{2}, \ldots, e_{i}\}$; otherwise ($i=0$), all $k$ experts are preserved.

\input{tables/ffn_sparse}

\section{Experiments}
\subsection{Performance of Sparse Attention}
We evaluate the performance of sparse attention verification with EAGLE-3 \citep{li2025eagle}.
We adopt Llama3.1-8B-Instruct \citep{dubey2024llama} as the target model.
Tree-structured drafts are used, which have 60 candidate tokens.
The experiment is conducted on 8 NVIDIA-H800-80G GPUs.
We evaluate the performance with sparse attention on 5 datasets in LongBench \citep{bai2024longbench} (including summarization, question answering, and code completion tasks): GovReport \citep{huang2021efficient}, 2WikiMQA \citep{ho2020constructing}, HotpotQA \citep{yang2018hotpotqa}, LCC \citep{guo2023longcoder}, RepoBench-P \citep{liu2024repobench}.
For sparse attention, the sparsity coefficient $\rho$ is set to 0.1.
To test the performance under different sparsity, we search the basic length $L_{0}$ in \{$1K, 2K, 4K$\}. We also compare the results with and without our proposed inter-layer retrieval reuse strategy.

Results are shown in Table \ref{tab:sparse_attn}.
Given a base length $L_{0}$, the sparsity ratio adapts based on the average sequence lengths across different datasets.
Under moderate sparsity levels (e.g., $L_{0}=4K$), SA exhibits minimal degradation compared with strict verification, with ROUGE/F1 drops typically within 0.3-1.0 points. This indicates that a large portion of attention computation is redundant for verification.
When sparsity increases, performance decreases more noticeably, yet the degradation remains within an acceptable range for most datasets. Datasets with stronger local dependencies (e.g., HotpotQA and LCC) show better robustness, suggesting that verification mainly relies on a subset of salient contextual tokens rather than the full attention map.

Introducing inter-layer reuse (SA*) slightly changes the trade-off. SA* consistently performs similarly to or slightly below SA, especially on QA datasets (e.g., 2WikiMQA, HotpotQA). This implies that while reuse reduces attention computation further, it introduces mild information loss due to cross-layer propagation of sparse patterns. However, for long-input summarization tasks such as GovReport and code-edit similarity tasks such as RepoBench-P, SA* performs comparably or even slightly better, indicating that reuse is more beneficial when long-range consistency, rather than fine-grained token interactions dominates the verification process.

\subsection{Performance of Sparse FFN}
\label{sec:sffn}

Similar to sparse attention, we also evaluate the performance of sparse FFN with EAGLE-3.
We employ Qwen3-30B-A3B \citep{qwen2025qwen25technicalreport} as the target model.
We apply SFFN to each expert.
Unlike SA, SFFN is also applicable to short texts. Therefore, in addition to the three long text datasets, we added additional mathematical tasks for our experiments, including GSM8K \citep{cobbe2021gsm8k}, Math \citep{hendrycks2021measuring}, and CollegeMath \citep{tang2024mathscalescalinginstructiontuning}.
Note that we remove LCC and RepoBench-P because Qwen3-MoE performs poorly on code tasks.
We search the threshold $\tau$ in \{$0.01, 0.05, 0.1$\} for channel selection.

Table \ref{tab:sparse_ffn} displays the results under different sparsity levels.
First, summarization (GovReport) remains stable across all sparsity levels. Notably, the ROUGE score slightly improves from 32.67 (strict) to 33.51 at $s_{f}=0.64$. This suggests that SFFN verification does not disrupt the model's semantic structure checking and may even introduce regularization effects beneficial for long-form text generation.
For QA (2WikiMQA, HotpotQA), performance also remains highly robust, indicating that reducing FFN computations in verification has minimal impact on tasks requiring compositional reasoning.
Regarding math reasoning tasks (GSM8K, Math, CollegeMath), sparse verification again shows strong resilience. Importantly, even at $s_{f}=0.64$, accuracy degradation is negligible, indicating that SFFN preserves the logical consistency required for step-by-step reasoning during verification.

\input{tables/sparse_moe}
\input{tables/h_sparse}
Across all benchmarks, we observe no systematic degradation as sparsity increases, even though up to nearly two-thirds of FFN activations are pruned. These results indicate that SFFN verification achieves a more computationally efficient mechanism without compromising the correctness.

\subsection{Performance of Sparse MoE}

We evaluate the performance of Sparse MoE with Deepseek-R1 \citep{guo2025deepseek} on the same datasets as in Section \ref{sec:sffn}.
Deepseek-R1 assigns 8 experts for each token in the MoE layers.
We consider a skipping budget $m$ in \{$2,3,4$\}.
We adopt the corresponding MTP (Multi-Token Prediction) heads as the draft model.
The draft length is set to 4.

Table \ref{tab:sparse_moe} shows that sparse MoE verification (SMoE) reduces expert usage while preserving competitive accuracy. Moderate sparsification ($m=2,3$) yields stable or even improved performance on several datasets (e.g., 2WikiMQA and Math with $m=3$), indicating that skipping low-contribution experts can remove noisy validation signals. However, as sparsity increases further ($m=4$), performance begins to degrade, particularly on math-heavy tasks, showing that excessive expert skipping may remove informative reasoning paths. Therefore, SMoE requires careful sparsity control to avoid harming verification reliability.

\subsection{Hybrid Sparse Methods}
While sparse attention, sparse FFN, and sparse MoE can independently reduce the verification cost of speculative decoding, each technique targets a different computational bottleneck and introduces distinct approximation errors. Therefore, it is crucial to evaluate whether combining multiple sparse mechanisms can achieve complementary benefits without causing systematic degradation in verification quality.
We evaluate the performance of a hybrid sparse verification method with the proposed strategies.
We use Deepseek-R1 as the target model and conduct speculative decoding with MTP.
The Basic length $L_{0}$ for SA is set to 4K.
Deepseek-R1 has 61 layers, of which 30 layers are selected as anchor layers to provide retrieval results to other layers.
The Threshold $\tau$ for FFN is set to 0.05.
The Expert skipping budget $m$ is set to 3.

Table \ref{tab:h_sparse} reports the performance of strict verification and the proposed hybrid method with three-dimensional sparsity. Compared with the strict baseline, the hybrid method achieves higher scores on GovReport, 2WikiMQA, and HotpotQA, while maintaining comparable performance on GSM8K and Math. These results indicate that the hybrid strategy does not lead to systematic degradation on long-context or multi-hop reasoning tasks under the evaluated sparsity configuration.
While combining sparsity across multiple dimensions yields additional efficiency gains, the resulting performance does not always improve monotonically. This indicates that sparsity dimensions are not strictly independent and should be coordinated rather than applied aggressively in isolation.

At the same time, performance drops are observed on CollegeMath, suggesting that tasks requiring fine-grained symbolic reasoning are more sensitive to aggressive sparsification during verification.
We observe that the impact of sparse verification varies across task types. Generation-oriented tasks such as summarization and open-domain QA exhibit relatively stable performance under moderate sparsity, whereas reasoning-intensive tasks, particularly mathematical problem solving, are more sensitive to sparsification. This suggests that tasks requiring precise token-level verification may demand stricter alignment between draft and target models.

\subsection{Impact on Acceptance Length}
Besides computational cost, the mean acceptance length is a crucial factor affecting the efficiency of speculative decoding. Therefore, we discuss the impact of these sparse verification methods on the average acceptance length in this section.

We report the mean acceptance length on each dataset in Table \ref{tab:h_sparse}.
The acceptance length is primarily determined by the alignment between the draft and target models. Since the draft model is trained to align with a non-sparse target model, sparsification in the target model can negatively affect this alignment.
While sparsification effectively reduces verification cost, it also introduces distributional shifts between the draft and target models, which can negatively affect token acceptance.
The mean acceptance length $\alpha$ shows a consistent but negligible reduction across most datasets, which has little effect on inference efficiency.

\section{Related Work}
\textbf{Sparse Inference} has been explored in multiple dimensions of language models.
KV Cache Eviction \citep{zhang2023h2o, ge2024model,xiaoefficient,tang2024quest,li2024snapkv} has emerged as an effective approach to mitigate the memory and latency overhead of large language model inference as the context length continues to grow.
$H_{2}O$ \citep{zhang2023h2o} proposes an oracle-based method that retains heavy-hitter tokens with the highest attention importance.
StreamingLLM \citep{xiaoefficient} reveals an attention sink phenomenon, where early key-value pairs consistently attract attention regardless of content, and leverages these stable positions along with recent context to maintain performance under limited cache.
Quest \citep{tang2024quest} introduces a query-aware mechanism that dynamically selects important tokens during decoding to reduce redundant cache entries.
SnapKV \citep{li2024snapkv} leverages attention observations at the head level to identify and preserve critical tokens, enabling fine-grained KV cache compression.
NSA \citep{yuan2025native} designs hardware-aligned, natively trainable sparse attention patterns to reduce computation and memory costs while maintaining model accuracy.

Sparse FFN methods focus on reducing computation in feed-forward layers. Early approaches \citep{roller2021hash,liu2023towards} exploit structured sparsity or conditional computation to activate only a subset of parameters for each token.
Mixture-of-Experts (MoE) models \citep{shazeer2017outrageouslylargeneuralnetworks, fedus2022switchtransformersscalingtrillion} route tokens to a small number of experts, significantly reducing per-token FLOPs while maintaining model capacity. Subsequent works further improve efficiency by refining routing strategies and expert utilization, including expert pruning, load balancing, and dynamic expert selection during inference \citep{rajbhandari2022deepspeedmoeadvancingmixtureofexpertsinference, lu-etal-2024-experts}.

\textbf{Speculative Decoding} \citep{stern2018blockwise,leviathan2023fast,xia2023speculative,xia-etal-2024-unlocking} follows a draft-verify paradigm to achieve lossless acceleration in autoregressive generation.
At each decoding step, a lightweight drafter first drafts several candidate tokens, which are then verified in parallel by the target model. The distribution of accepted tokens is consistent with autoregressive generation, allowing multiple tokens to be generated in a single step.
Some studies \citep{leviathan2023fast,cai2024medusa,li2024eagle} employ independently trained small models or auxiliary modules as draft generators, while others \citep{saxena2023prompt,fu2024break,he-etal-2024-rest} retrieve drafts from pre-constructed draft pools.
The draft structures have evolved from simple n-grams \citep{leviathan2023fast,fu2024break} to more sophisticated draft trees \citep{li2024eagle,wang2024opt}. 
Increasing the draft length imposes additional computational burden on the verification stage, particularly under high-load conditions such as long-context scenarios.
While prior studies \citep{sadhukhanmagicdec,yang2025longspec} have investigated KV cache compression in the draft stage for long contexts, the target model’s KV cache during verification is much larger and thus represents the key bottleneck.

\section{Conclusion}

This work investigates sparse verification in speculative decoding and demonstrates that substantial computational redundancy exists in the verification stage across multiple dimensions, including attention, feed-forward networks, and expert activation. By analyzing the structural characteristics of speculative verification, we show that sparsification strategies originally designed for standard decoding can be adapted to verification with careful control of acceptance behavior.
We propose a unified framework that integrates multi-dimensional sparsity into speculative verification without requiring additional training or model modification. In particular, the framework incorporates adaptive expert skipping and retrieval reuse to reduce redundant computation while preserving verification correctness. Empirical results across diverse benchmarks indicate that sparse verification achieves consistent efficiency gains, with only a moderate impact on acceptance length under appropriate sparsity levels.

\bibliography{custom}




\end{document}

%% file: tables/flops.tex
\begin{table}[t]
\centering
\resizebox{\columnwidth}{!}{
\begin{tabular}{@{}l l@{}}
\toprule
Module & FLOPs \\ \midrule

Full Attention & $6\, B T d^{2} \;+\; 4\, B T^{2} d$ \\
Sparse Attention & $6\, B T d^{2} \;+\; 4\, B \, (1-s_{a}) \, T^{2} d$ \\
\hline
Dense FFN & $6\, B T d\, d_{\mathrm{f}}$ \\
Sparse FFN & $2 B T d\, d_{\mathrm{f}} \,(3 - 2s_f)$\\ 
\hline
MoE FFN  & $4\, B T\, k\, d\, d_{\mathrm{e}} \;+\; 2\, B T d\, E$ \\
Sparse MoE & $4\, B T\, (1-s_{e})k\, d\, d_{\mathrm{e}} \;+\; 2\, B T d\, E$\\

\bottomrule
\end{tabular}}
\caption{\label{tab:flops} FLOPs estimation for different modules. $L$: number of layers, $d$: hidden size, $d_{\mathrm{f}}$: FFN intermediate size, $d_{\mathrm{e}}$: expert hidden size, $n_h$: number of attention heads, $T$: sequence length, $B$: batch size, $E$: number of experts, $k$: number of active experts per token. $s_{a},s_f,s_{e}$ indicate sparsity of attention, sparsity of down projection in FFN, sparsity of active experts.}
\end{table}

%% file: tables/sparse_attn.tex
\begin{table*}[t]
\centering
\resizebox{\textwidth}{!}{
\begin{tabularx}{\textwidth}{l*{10}{>{\centering\arraybackslash}X}}
\toprule
\multirow{2}{*}{\textbf{Verification}}      & \multicolumn{2}{c}{\textbf{GovReport}}  & \multicolumn{2}{c}{\textbf{2WikiMQA}} & \multicolumn{2}{c}{\textbf{HotpotQA}} & \multicolumn{2}{c}{\textbf{LCC}} & \multicolumn{2}{c}{\textbf{RepoBench-P}}  \\
\cmidrule{2-11}
 & $s_{a}$ & {\scriptsize ROUGE} & $s_{a}$ & F1 &  $s_{a}$ & F1 & $s_{a}$ & {\scriptsize Edit Sim} & $s_{a}$ & {\scriptsize Edit Sim}\\
\midrule
\rowcolor{gray!35}
Strict           &   0 & 34.18 & 0  & 39.20 & 0 & 47.66 & 0 & 54.24 & 0 & 21.76        \\
SA ($L_{0}=4K$) &    \multirow{2}{*}{0.34}   &  33.50        &   \multirow{2}{*}{0.39}       &  39.32   & \multirow{2}{*}{0.41}  & 47.43 & \multirow{2}{*}{0.41} & 55.10 & \multirow{2}{*}{0.45} & 22.23 \\  
SA* ($L_{0}=4K$) &           &    33.28     &          &   35.78  &  & 44.78 & & 55.18 & & 22.24\\ 
SA ($L_{0}=2K$) &    \multirow{2}{*}{0.60}   &     31.86     &    \multirow{2}{*}{0.63}      &   38.81  & \multirow{2}{*}{0.64} &  46.55 & \multirow{2}{*}{0.56} & 55.24  & \multirow{2}{*}{0.66}  & 22.00 \\  
SA* ($L_{0}=2K$) &           &   31.67       &          &   36.77  &    & 44.18 & & 54.27 & & 22.14\\ 
SA ($L_{0}=1K$) &  \multirow{2}{*}{0.75}     &    30.62      &     \multirow{2}{*}{0.77}     &   38.86  & \multirow{2}{*}{0.77} & 46.28  &  \multirow{2}{*}{0.72} & 54.99 & \multirow{2}{*}{0.78} & 21.61 \\  
SA* ($L_{0}=1K$) &           &    29.45      &          &  37.87   &    & 43.18 & & 53.52& & 21.76 \\  
\bottomrule
\end{tabularx}}
\caption{\label{tab:sparse_attn} Performance of verification with sparse attention on different datasets. ``Strict'' indicates standard speculative decoding with strict verification with full attention. ``SA'' represents sparse attention without inter-layer retrieval reuse strategy, while ``SA*'' represents sparse attention with the reuse strategy. $s_{a}$ is sparsity of attention.}
\end{table*}

%% file: tables/ffn_sparse.tex
\begin{table*}[t]
\centering
\resizebox{\textwidth}{!}{
\begin{tabularx}{\textwidth}{l*{8}{>{\centering\arraybackslash}X}}
\toprule
\multirow{2}{*}{\textbf{Verification}}  & \multirow{2}{*}{$s_{f}$}    & \textbf{\small GovReport}  & \textbf{\small 2WikiMQA} & \textbf{\small HotpotQA} & \textbf{\small GSM8K} & \textbf{\small Math} & \textbf{\small CollegeMath} \\
\cmidrule{3-8}
 & & {\footnotesize ROUGE} &  F1 &  F1 & Acc. & Acc. & Acc. \\
\midrule
\rowcolor{gray!35}
Strict     &     0    & 32.67 & 43.93 & 63.32 & 90.0 & 65.6 & 23.8\\
SFFN ($\tau=0.01$)        &     0.18   & 32.40 & 46.77 & 63.36 & 91.0 & 63.0 & 23.6\\
SFFN ($\tau=0.05$)      &     0.47   & 32.96 & 43.99 & 63.33 & 90.7 & 63.0 & 23.8\\
SFFN ($\tau=0.1$)         &     0.64  &  33.51 & 41.96 & 62.01 & 91.0 & 63.6 & 24.8 \\

\bottomrule
\end{tabularx}}
\caption{\label{tab:sparse_ffn} Performance of verification with sparse FFN on different datasets with Qwen3-30B-A3B. ``Strict'' indicates standard speculative decoding with strict verification. $s_{f}$ is the mean sparsity of FFN module.}
\end{table*}

%% file: tables/sparse_moe.tex
\begin{table*}[t]
\centering
\resizebox{\textwidth}{!}{
\begin{tabularx}{\textwidth}{l*{8}{>{\centering\arraybackslash}X}}
\toprule
\multirow{2}{*}{\textbf{Verification}}  & \multirow{2}{*}{$s_{e}$}    & \textbf{\small GovReport}  & \textbf{\small 2WikiMQA} & \textbf{\small HotpotQA} & \textbf{\small GSM8K} & \textbf{\small Math} & \textbf{\small CollegeMath}  \\
\cmidrule{3-8}
 & & {\footnotesize ROUGE} &  F1 &  F1 & Acc. & Acc. & Acc. \\
\midrule
\rowcolor{gray!35}
Strict     &     0    & 26.90 & 77.39 & 73.43 & 98.0 & 82.0 & 58.0\\
SMoE ($m=2$)    &    0.11   & 26.74 & 79.81 & 74.60 & 98.0 & 80.0 & 56.0 \\
SMoE ($m=3$)  &    0.16   & 26.73 & 78.52 & 75.03 & 97.0 & 87.0 & 57.0 \\
SMoE ($m=4$)      &    0.22  &  26.76 & 78.88 & 74.77 & 95.0 & 75.0 & 52.0 \\

\bottomrule
\end{tabularx}}
\caption{\label{tab:sparse_moe} Performance of verification with sparse MoE on different datasets with Deepseek-R1. ``Strict'' indicates standard speculative decoding with strict verification. $s_{e}$ is the mean sparsity of activated experts.}
\end{table*}

%% file: tables/h_sparse.tex
\begin{table*}[t]
\centering
\resizebox{\textwidth}{!}{
\begin{tabularx}{\textwidth}{l*{13}{>{\centering\arraybackslash}X}}
\toprule
\multirow{2}{*}{\textbf{Verification}}      & \multicolumn{2}{c}{\textbf{\small GovReport}}  & \multicolumn{2}{c}{\textbf{\small 2WikiMQA}} & \multicolumn{2}{c}{\textbf{\small HotpotQA}} & \multicolumn{2}{c}{\textbf{\small GSM8K}} & \multicolumn{2}{c}{\textbf{\small Math}} & \multicolumn{2}{c}{\textbf{\small CollegeMath}}  \\
\cmidrule{2-13}
 & {\scriptsize ROUGE} & $\alpha$ & {\footnotesize F1} & $\alpha$ &  {\footnotesize F1} & $\alpha$ & {\footnotesize Acc.} & $\alpha$ & {\footnotesize Acc.} & $\alpha$ & {\footnotesize Acc.} & $\alpha$  \\
\midrule
\rowcolor{gray!35}
Strict        & 26.90 &2.44& 77.39 &2.71& 73.43 &2.67 & 98.0 &2.84& 82.0 &2.85& 58.0&2.89\\
Hybrid       & 27.40 & 2.42 & 79.82 & 2.70 & 74.57 & 2.66 & 96.0 & 2.81 & 81.0 &2.84&53.0 & 2.89 \\
\bottomrule
\end{tabularx}}
\caption{\label{tab:h_sparse} Performance of verification with hybrid method with the 3-dimensional sparsity on different datasets with Deepseek-R1. ``Strict'' indicates standard speculative decoding with strict verification. ``$\alpha$'' represents mean acceptance length.}
\end{table*}

%% file: custom.bib
@misc{qwen2025qwen25technicalreport,
      title={Qwen2.5 Technical Report}, 
      author={Qwen and : and An Yang and Baosong Yang and Beichen Zhang and Binyuan Hui and Bo Zheng and Bowen Yu and Chengyuan Li and Dayiheng Liu and Fei Huang and Haoran Wei and Huan Lin and Jian Yang and Jianhong Tu and Jianwei Zhang and Jianxin Yang and Jiaxi Yang and Jingren Zhou and Junyang Lin and Kai Dang and Keming Lu and Keqin Bao and Kexin Yang and Le Yu and Mei Li and Mingfeng Xue and Pei Zhang and Qin Zhu and Rui Men and Runji Lin and Tianhao Li and Tianyi Tang and Tingyu Xia and Xingzhang Ren and Xuancheng Ren and Yang Fan and Yang Su and Yichang Zhang and Yu Wan and Yuqiong Liu and Zeyu Cui and Zhenru Zhang and Zihan Qiu},
      year={2025},
      eprint={2412.15115},
      archivePrefix={arXiv},
      primaryClass={cs.CL},
      url={https://arxiv.org/abs/2412.15115}, 
}

@inproceedings{leviathan2023fast,
  title={Fast inference from transformers via speculative decoding},
  author={Leviathan, Yaniv and Kalman, Matan and Matias, Yossi},
  booktitle={International Conference on Machine Learning},
  pages={19274--19286},
  year={2023},
  organization={PMLR}
}

@article{stern2018blockwise,
  title={Blockwise parallel decoding for deep autoregressive models},
  author={Stern, Mitchell and Shazeer, Noam and Uszkoreit, Jakob},
  journal={Advances in Neural Information Processing Systems},
  volume={31},
  year={2018}
}

@inproceedings{xia2023speculative,
  title={Speculative decoding: Exploiting speculative execution for accelerating seq2seq generation},
  author={Xia, Heming and Ge, Tao and Wang, Peiyi and Chen, Si-Qing and Wei, Furu and Sui, Zhifang},
  booktitle={Findings of the Association for Computational Linguistics: EMNLP 2023},
  pages={3909--3925},
  year={2023}
}

@inproceedings{
cai2024medusa,
title={Medusa: Simple {LLM} Inference Acceleration Framework with Multiple Decoding Heads},
author={Tianle Cai and Yuhong Li and Zhengyang Geng and Hongwu Peng and Jason D. Lee and Deming Chen and Tri Dao},
booktitle={Forty-first International Conference on Machine Learning},
year={2024},
url={https://openreview.net/forum?id=PEpbUobfJv}
}

@inproceedings{
li2024eagle,
title={{EAGLE}: Speculative Sampling Requires Rethinking Feature Uncertainty},
author={Yuhui Li and Fangyun Wei and Chao Zhang and Hongyang Zhang},
booktitle={Forty-first International Conference on Machine Learning},
year={2024},
url={https://openreview.net/forum?id=1NdN7eXyb4}
}

@article{wang2024opt,
    author = {Wang, Jikai and Su, Yi and Li, Juntao and Xia, Qingrong and Ye, Zi and Duan, Xinyu and Wang, Zhefeng and Zhang, Min},
    title = {OPT-Tree: Speculative Decoding with Adaptive Draft Tree Structure},
    journal = {Transactions of the Association for Computational Linguistics},
    volume = {13},
    pages = {188-199},
    year = {2024},
    month = {02},
    issn = {2307-387X},
    doi = {10.1162/tacl_a_00735},
    url = {https://doi.org/10.1162/tacl\_a\_00735},
    eprint = {https://direct.mit.edu/tacl/article-pdf/doi/10.1162/tacl\_a\_00735/2506509/tacl\_a\_00735.pdf},
}

@misc{saxena2023prompt,
    title = {Prompt Lookup Decoding},
    author = {Apoorv Saxena},
    year = {2023},
    month = {November},
    url = {https://github.com/apoorvumang/prompt-lookup-decoding/}
}

@inproceedings{
fu2024break,
title={Break the Sequential Dependency of {LLM} Inference Using Lookahead Decoding},
author={Yichao Fu and Peter Bailis and Ion Stoica and Hao Zhang},
booktitle={Forty-first International Conference on Machine Learning},
year={2024},
url={https://openreview.net/forum?id=eDjvSFOkXw}
}

@inproceedings{he-etal-2024-rest,
    title = "{REST}: Retrieval-Based Speculative Decoding",
    author = "He, Zhenyu  and
      Zhong, Zexuan  and
      Cai, Tianle  and
      Lee, Jason  and
      He, Di",
    editor = "Duh, Kevin  and
      Gomez, Helena  and
      Bethard, Steven",
    booktitle = "Proceedings of the 2024 Conference of the North American Chapter of the Association for Computational Linguistics: Human Language Technologies (Volume 1: Long Papers)",
    month = jun,
    year = "2024",
    address = "Mexico City, Mexico",
    publisher = "Association for Computational Linguistics",
    url = "https://aclanthology.org/2024.naacl-long.88",
    doi = "10.18653/v1/2024.naacl-long.88",
    pages = "1582--1595",
}

@inproceedings{ho2020constructing,
  title={Constructing A Multi-hop QA Dataset for Comprehensive Evaluation of Reasoning Steps},
  author={Ho, Xanh and Nguyen, Anh-Khoa Duong and Sugawara, Saku and Aizawa, Akiko},
  booktitle={Proceedings of the 28th International Conference on Computational Linguistics},
  pages={6609--6625},
  year={2020}
}

@inproceedings{yang2018hotpotqa,
  title={HotpotQA: A Dataset for Diverse, Explainable Multi-hop Question Answering},
  author={Yang, Zhilin and Qi, Peng and Zhang, Saizheng and Bengio, Yoshua and Cohen, William and Salakhutdinov, Ruslan and Manning, Christopher D},
  booktitle={Proceedings of the 2018 Conference on Empirical Methods in Natural Language Processing},
  pages={2369--2380},
  year={2018}
}

@inproceedings{
liu2024repobench,
title={RepoBench: Benchmarking Repository-Level Code Auto-Completion Systems},
author={Tianyang Liu and Canwen Xu and Julian McAuley},
booktitle={The Twelfth International Conference on Learning Representations},
year={2024},
url={https://openreview.net/forum?id=pPjZIOuQuF}
}

@inproceedings{guo2023longcoder,
  title={Longcoder: A long-range pre-trained language model for code completion},
  author={Guo, Daya and Xu, Canwen and Duan, Nan and Yin, Jian and McAuley, Julian},
  booktitle={International Conference on Machine Learning},
  pages={12098--12107},
  year={2023},
  organization={PMLR}
}

@inproceedings{xia-etal-2024-unlocking,
    title = "Unlocking Efficiency in Large Language Model Inference: A Comprehensive Survey of Speculative Decoding",
    author = "Xia, Heming  and
      Yang, Zhe  and
      Dong, Qingxiu  and
      Wang, Peiyi  and
      Li, Yongqi  and
      Ge, Tao  and
      Liu, Tianyu  and
      Li, Wenjie  and
      Sui, Zhifang",
    editor = "Ku, Lun-Wei  and
      Martins, Andre  and
      Srikumar, Vivek",
    booktitle = "Findings of the Association for Computational Linguistics ACL 2024",
    month = aug,
    year = "2024",
    address = "Bangkok, Thailand and virtual meeting",
    publisher = "Association for Computational Linguistics",
    url = "https://aclanthology.org/2024.findings-acl.456",
    pages = "7655--7671",
}

@misc{openai2024gpt4technicalreport,
      title={GPT-4 Technical Report}, 
      author={OpenAI and Josh Achiam and Steven Adler and Sandhini Agarwal and Lama Ahmad and Ilge Akkaya and Florencia Leoni Aleman and others},
      year={2024},
      eprint={2303.08774},
      archivePrefix={arXiv},
      primaryClass={cs.CL},
      url={https://arxiv.org/abs/2303.08774}, 
}

@misc{
llama31,
author = {Meta-AI},
title={Introducing Llama 3.1: Our most capable models to date},
year = {2024},
url={https://ai.meta.com/blog/meta-llama-3-1/}
}

@inproceedings{sadhukhanmagicdec,
  title={MagicDec: Breaking the Latency-Throughput Tradeoff for Long Context Generation with Speculative Decoding},
  author={Sadhukhan, Ranajoy and Chen, Jian and Chen, Zhuoming and Tiwari, Vashisth and Lai, Ruihang and Shi, Jinyuan and Yen, Ian En-Hsu and May, Avner and Chen, Tianqi and Chen, Beidi},
  booktitle={The Thirteenth International Conference on Learning Representations},
  year={2024}
}

@article{yang2025longspec,
  title={LongSpec: Long-Context Speculative Decoding with Efficient Drafting and Verification},
  author={Yang, Penghui and Du, Cunxiao and Zhang, Fengzhuo and Wang, Haonan and Pang, Tianyu and Du, Chao and An, Bo},
  journal={arXiv preprint arXiv:2502.17421},
  year={2025}
}

@inproceedings{tang2024quest,
  title={QUEST: Query-Aware Sparsity for Efficient Long-Context LLM Inference},
  author={Tang, Jiaming and Zhao, Yilong and Zhu, Kan and Xiao, Guangxuan and Kasikci, Baris and Han, Song},
  booktitle={International Conference on Machine Learning},
  pages={47901--47911},
  year={2024},
  organization={PMLR}
}

@inproceedings{xiaoefficient,
  title={Efficient Streaming Language Models with Attention Sinks},
  author={Xiao, Guangxuan and Tian, Yuandong and Chen, Beidi and Han, Song and Lewis, Mike},
  booktitle={The Twelfth International Conference on Learning Representations},
    year={2024},
}

@inproceedings{yang2018modeling,
  title={Modeling Localness for Self-Attention Networks},
  author={Yang, Baosong and Tu, Zhaopeng and Wong, Derek F and Meng, Fandong and Chao, Lidia S and Zhang, Tong},
  booktitle={Proceedings of the 2018 Conference on Empirical Methods in Natural Language Processing},
  pages={4449--4458},
  year={2018}
}

@inproceedings{wuemergence,
  title={On the Emergence of Position Bias in Transformers},
  author={Wu, Xinyi and Wang, Yifei and Jegelka, Stefanie and Jadbabaie, Ali},
  booktitle={Forty-second International Conference on Machine Learning},
year={2025}
}

@inproceedings{ge2024model,
  title={MODEL TELLS YOU WHAT TO DISCARD: ADAPTIVE KV CACHE COMPRESSION FOR LLMS},
  author={Ge, Suyu and Zhang, Yunan and Liu, Liyuan and Zhang, Minjia and Han, Jiawei and Gao, Jianfeng},
  booktitle={12th International Conference on Learning Representations, ICLR 2024},
  year={2024}
}

@article{li2024snapkv,
  title={Snapkv: Llm knows what you are looking for before generation},
  author={Li, Yuhong and Huang, Yingbing and Yang, Bowen and Venkitesh, Bharat and Locatelli, Acyr and Ye, Hanchen and Cai, Tianle and Lewis, Patrick and Chen, Deming},
  journal={Advances in Neural Information Processing Systems},
  volume={37},
  pages={22947--22970},
  year={2024}
}

@article{zhang2023h2o,
  title={H2o: Heavy-hitter oracle for efficient generative inference of large language models},
  author={Zhang, Zhenyu and Sheng, Ying and Zhou, Tianyi and Chen, Tianlong and Zheng, Lianmin and Cai, Ruisi and Song, Zhao and Tian, Yuandong and R{\'e}, Christopher and Barrett, Clark and others},
  journal={Advances in Neural Information Processing Systems},
  volume={36},
  pages={34661--34710},
  year={2023}
}

@article{yuan2025native,
  title={Native sparse attention: Hardware-aligned and natively trainable sparse attention},
  author={Yuan, Jingyang and Gao, Huazuo and Dai, Damai and Luo, Junyu and Zhao, Liang and Zhang, Zhengyan and Xie, Zhenda and Wei, YX and Wang, Lean and Xiao, Zhiping and others},
  journal={arXiv preprint arXiv:2502.11089},
  year={2025}
}

@misc{lu2025mobamixtureblockattention,
      title={MoBA: Mixture of Block Attention for Long-Context LLMs}, 
      author={Enzhe Lu and Zhejun Jiang and Jingyuan Liu and Yulun Du and Tao Jiang and Chao Hong and Shaowei Liu and Weiran He and Enming Yuan and Yuzhi Wang and Zhiqi Huang and Huan Yuan and Suting Xu and Xinran Xu and Guokun Lai and Yanru Chen and Huabin Zheng and Junjie Yan and Jianlin Su and Yuxin Wu and Neo Y. Zhang and Zhilin Yang and Xinyu Zhou and Mingxing Zhang and Jiezhong Qiu},
      year={2025},
      eprint={2502.13189},
      archivePrefix={arXiv},
      primaryClass={cs.LG},
      url={https://arxiv.org/abs/2502.13189}, 
}

@misc{grattafiori2024llama3herdmodels,
      title={The Llama 3 Herd of Models}, 
      author={Aaron Grattafiori and Abhimanyu Dubey and Abhinav Jauhri and Abhinav Pandey and Abhishek Kadian and Ahmad Al-Dahle and Aiesha Letman and Akhil Mathur and Alan Schelten and Alex Vaughan and Amy Yang and Angela Fan and Anirudh Goyal and Anthony Hartshorn and Aobo Yang and Archi Mitra and Archie Sravankumar and Artem Korenev and Arthur Hinsvark and Arun Rao and Aston Zhang and Aurelien Rodriguez and Austen Gregerson and Ava Spataru and Baptiste Roziere and Bethany Biron and Binh Tang and Bobbie Chern and Charlotte Caucheteux and Chaya Nayak and Chloe Bi and Chris Marra and Chris McConnell and Christian Keller and Christophe Touret and Chunyang Wu and Corinne Wong and Cristian Canton Ferrer and Cyrus Nikolaidis and Damien Allonsius and Daniel Song and Danielle Pintz and Danny Livshits and Danny Wyatt and David Esiobu and Dhruv Choudhary and Dhruv Mahajan and Diego Garcia-Olano and Diego Perino and Dieuwke Hupkes and Egor Lakomkin and Ehab AlBadawy and Elina Lobanova and Emily Dinan and Eric Michael Smith and Filip Radenovic and Francisco Guzmán and Frank Zhang and Gabriel Synnaeve and Gabrielle Lee and Georgia Lewis Anderson and Govind Thattai and Graeme Nail and Gregoire Mialon and Guan Pang and Guillem Cucurell and Hailey Nguyen and Hannah Korevaar and Hu Xu and Hugo Touvron and Iliyan Zarov and Imanol Arrieta Ibarra and Isabel Kloumann and Ishan Misra and Ivan Evtimov and Jack Zhang and Jade Copet and Jaewon Lee and Jan Geffert and Jana Vranes and Jason Park and Jay Mahadeokar and Jeet Shah and Jelmer van der Linde and Jennifer Billock and Jenny Hong and Jenya Lee and Jeremy Fu and Jianfeng Chi and Jianyu Huang and Jiawen Liu and Jie Wang and Jiecao Yu and Joanna Bitton and Joe Spisak and Jongsoo Park and Joseph Rocca and Joshua Johnstun and Joshua Saxe and Junteng Jia and Kalyan Vasuden Alwala and Karthik Prasad and Kartikeya Upasani and Kate Plawiak and Ke Li and Kenneth Heafield and Kevin Stone and Khalid El-Arini and Krithika Iyer and Kshitiz Malik and Kuenley Chiu and Kunal Bhalla and Kushal Lakhotia and Lauren Rantala-Yeary and Laurens van der Maaten and Lawrence Chen and Liang Tan and Liz Jenkins and Louis Martin and Lovish Madaan and Lubo Malo and Lukas Blecher and Lukas Landzaat and Luke de Oliveira and Madeline Muzzi and Mahesh Pasupuleti and Mannat Singh and Manohar Paluri and Marcin Kardas and Maria Tsimpoukelli and Mathew Oldham and Mathieu Rita and Maya Pavlova and Melanie Kambadur and Mike Lewis and Min Si and Mitesh Kumar Singh and Mona Hassan and Naman Goyal and Narjes Torabi and Nikolay Bashlykov and Nikolay Bogoychev and Niladri Chatterji and Ning Zhang and Olivier Duchenne and Onur Çelebi and Patrick Alrassy and Pengchuan Zhang and Pengwei Li and Petar Vasic and Peter Weng and Prajjwal Bhargava and Pratik Dubal and Praveen Krishnan and Punit Singh Koura and Puxin Xu and Qing He and Qingxiao Dong and Ragavan Srinivasan and Raj Ganapathy and Ramon Calderer and Ricardo Silveira Cabral and Robert Stojnic and Roberta Raileanu and Rohan Maheswari and Rohit Girdhar and Rohit Patel and Romain Sauvestre and Ronnie Polidoro and Roshan Sumbaly and Ross Taylor and Ruan Silva and Rui Hou and Rui Wang and Saghar Hosseini and Sahana Chennabasappa and Sanjay Singh and Sean Bell and Seohyun Sonia Kim and Sergey Edunov and Shaoliang Nie and Sharan Narang and Sharath Raparthy and Sheng Shen and Shengye Wan and Shruti Bhosale and Shun Zhang and Simon Vandenhende and Soumya Batra and Spencer Whitman and Sten Sootla and Stephane Collot and Suchin Gururangan and Sydney Borodinsky and Tamar Herman and Tara Fowler and Tarek Sheasha and Thomas Georgiou and Thomas Scialom and Tobias Speckbacher and Todor Mihaylov and Tong Xiao and Ujjwal Karn and Vedanuj Goswami and Vibhor Gupta and Vignesh Ramanathan and Viktor Kerkez and Vincent Gonguet and Virginie Do and Vish Vogeti and Vítor Albiero and Vladan Petrovic and Weiwei Chu and Wenhan Xiong and Wenyin Fu and Whitney Meers and Xavier Martinet and Xiaodong Wang and Xiaofang Wang and Xiaoqing Ellen Tan and Xide Xia and Xinfeng Xie and Xuchao Jia and Xuewei Wang and Yaelle Goldschlag and Yashesh Gaur and Yasmine Babaei and Yi Wen and Yiwen Song and Yuchen Zhang and Yue Li and Yuning Mao and Zacharie Delpierre Coudert and Zheng Yan and Zhengxing Chen and Zoe Papakipos and Aaditya Singh and Aayushi Srivastava and Abha Jain and Adam Kelsey and Adam Shajnfeld and Adithya Gangidi and Adolfo Victoria and Ahuva Goldstand and Ajay Menon and Ajay Sharma and Alex Boesenberg and Alexei Baevski and Allie Feinstein and Amanda Kallet and Amit Sangani and Amos Teo and Anam Yunus and Andrei Lupu and Andres Alvarado and Andrew Caples and Andrew Gu and Andrew Ho and Andrew Poulton and Andrew Ryan and Ankit Ramchandani and Annie Dong and Annie Franco and Anuj Goyal and Aparajita Saraf and Arkabandhu Chowdhury and Ashley Gabriel and Ashwin Bharambe and Assaf Eisenman and Azadeh Yazdan and Beau James and Ben Maurer and Benjamin Leonhardi and Bernie Huang and Beth Loyd and Beto De Paola and Bhargavi Paranjape and Bing Liu and Bo Wu and Boyu Ni and Braden Hancock and Bram Wasti and Brandon Spence and Brani Stojkovic and Brian Gamido and Britt Montalvo and Carl Parker and Carly Burton and Catalina Mejia and Ce Liu and Changhan Wang and Changkyu Kim and Chao Zhou and Chester Hu and Ching-Hsiang Chu and Chris Cai and Chris Tindal and Christoph Feichtenhofer and Cynthia Gao and Damon Civin and Dana Beaty and Daniel Kreymer and Daniel Li and David Adkins and David Xu and Davide Testuggine and Delia David and Devi Parikh and Diana Liskovich and Didem Foss and Dingkang Wang and Duc Le and Dustin Holland and Edward Dowling and Eissa Jamil and Elaine Montgomery and Eleonora Presani and Emily Hahn and Emily Wood and Eric-Tuan Le and Erik Brinkman and Esteban Arcaute and Evan Dunbar and Evan Smothers and Fei Sun and Felix Kreuk and Feng Tian and Filippos Kokkinos and Firat Ozgenel and Francesco Caggioni and Frank Kanayet and Frank Seide and Gabriela Medina Florez and Gabriella Schwarz and Gada Badeer and Georgia Swee and Gil Halpern and Grant Herman and Grigory Sizov and Guangyi and Zhang and Guna Lakshminarayanan and Hakan Inan and Hamid Shojanazeri and Han Zou and Hannah Wang and Hanwen Zha and Haroun Habeeb and Harrison Rudolph and Helen Suk and Henry Aspegren and Hunter Goldman and Hongyuan Zhan and Ibrahim Damlaj and Igor Molybog and Igor Tufanov and Ilias Leontiadis and Irina-Elena Veliche and Itai Gat and Jake Weissman and James Geboski and James Kohli and Janice Lam and Japhet Asher and Jean-Baptiste Gaya and Jeff Marcus and Jeff Tang and Jennifer Chan and Jenny Zhen and Jeremy Reizenstein and Jeremy Teboul and Jessica Zhong and Jian Jin and Jingyi Yang and Joe Cummings and Jon Carvill and Jon Shepard and Jonathan McPhie and Jonathan Torres and Josh Ginsburg and Junjie Wang and Kai Wu and Kam Hou U and Karan Saxena and Kartikay Khandelwal and Katayoun Zand and Kathy Matosich and Kaushik Veeraraghavan and Kelly Michelena and Keqian Li and Kiran Jagadeesh and Kun Huang and Kunal Chawla and Kyle Huang and Lailin Chen and Lakshya Garg and Lavender A and Leandro Silva and Lee Bell and Lei Zhang and Liangpeng Guo and Licheng Yu and Liron Moshkovich and Luca Wehrstedt and Madian Khabsa and Manav Avalani and Manish Bhatt and Martynas Mankus and Matan Hasson and Matthew Lennie and Matthias Reso and Maxim Groshev and Maxim Naumov and Maya Lathi and Meghan Keneally and Miao Liu and Michael L. Seltzer and Michal Valko and Michelle Restrepo and Mihir Patel and Mik Vyatskov and Mikayel Samvelyan and Mike Clark and Mike Macey and Mike Wang and Miquel Jubert Hermoso and Mo Metanat and Mohammad Rastegari and Munish Bansal and Nandhini Santhanam and Natascha Parks and Natasha White and Navyata Bawa and Nayan Singhal and Nick Egebo and Nicolas Usunier and Nikhil Mehta and Nikolay Pavlovich Laptev and Ning Dong and Norman Cheng and Oleg Chernoguz and Olivia Hart and Omkar Salpekar and Ozlem Kalinli and Parkin Kent and Parth Parekh and Paul Saab and Pavan Balaji and Pedro Rittner and Philip Bontrager and Pierre Roux and Piotr Dollar and Polina Zvyagina and Prashant Ratanchandani and Pritish Yuvraj and Qian Liang and Rachad Alao and Rachel Rodriguez and Rafi Ayub and Raghotham Murthy and Raghu Nayani and Rahul Mitra and Rangaprabhu Parthasarathy and Raymond Li and Rebekkah Hogan and Robin Battey and Rocky Wang and Russ Howes and Ruty Rinott and Sachin Mehta and Sachin Siby and Sai Jayesh Bondu and Samyak Datta and Sara Chugh and Sara Hunt and Sargun Dhillon and Sasha Sidorov and Satadru Pan and Saurabh Mahajan and Saurabh Verma and Seiji Yamamoto and Sharadh Ramaswamy and Shaun Lindsay and Shaun Lindsay and Sheng Feng and Shenghao Lin and Shengxin Cindy Zha and Shishir Patil and Shiva Shankar and Shuqiang Zhang and Shuqiang Zhang and Sinong Wang and Sneha Agarwal and Soji Sajuyigbe and Soumith Chintala and Stephanie Max and Stephen Chen and Steve Kehoe and Steve Satterfield and Sudarshan Govindaprasad and Sumit Gupta and Summer Deng and Sungmin Cho and Sunny Virk and Suraj Subramanian and Sy Choudhury and Sydney Goldman and Tal Remez and Tamar Glaser and Tamara Best and Thilo Koehler and Thomas Robinson and Tianhe Li and Tianjun Zhang and Tim Matthews and Timothy Chou and Tzook Shaked and Varun Vontimitta and Victoria Ajayi and Victoria Montanez and Vijai Mohan and Vinay Satish Kumar and Vishal Mangla and Vlad Ionescu and Vlad Poenaru and Vlad Tiberiu Mihailescu and Vladimir Ivanov and Wei Li and Wenchen Wang and Wenwen Jiang and Wes Bouaziz and Will Constable and Xiaocheng Tang and Xiaojian Wu and Xiaolan Wang and Xilun Wu and Xinbo Gao and Yaniv Kleinman and Yanjun Chen and Ye Hu and Ye Jia and Ye Qi and Yenda Li and Yilin Zhang and Ying Zhang and Yossi Adi and Youngjin Nam and Yu and Wang and Yu Zhao and Yuchen Hao and Yundi Qian and Yunlu Li and Yuzi He and Zach Rait and Zachary DeVito and Zef Rosnbrick and Zhaoduo Wen and Zhenyu Yang and Zhiwei Zhao and Zhiyu Ma},
      year={2024},
      eprint={2407.21783},
      archivePrefix={arXiv},
      primaryClass={cs.AI},
      url={https://arxiv.org/abs/2407.21783}, 
}

@inproceedings{liu2023towards,
  title={Towards a unified view of sparse feed-forward network in pretraining large language model},
  author={Liu, Zeyu and Dettmers, Tim and Lin, Xi and Stoyanov, Veselin and Li, Xian},
  booktitle={Proceedings of the 2023 Conference on Empirical Methods in Natural Language Processing},
  pages={15038--15061},
  year={2023}
}

@article{roller2021hash,
  title={Hash layers for large sparse models},
  author={Roller, Stephen and Sukhbaatar, Sainbayar and Weston, Jason and others},
  journal={advances in neural information processing systems},
  volume={34},
  pages={17555--17566},
  year={2021}
}

@article{hendrycks2016gaussian,
  title={Gaussian Error Linear Units (Gelus)},
  author={Hendrycks, D},
  journal={arXiv preprint arXiv:1606.08415},
  year={2016}
}

@article{shazeer2020glu,
  title={Glu variants improve transformer},
  author={Shazeer, Noam},
  journal={arXiv preprint arXiv:2002.05202},
  year={2020}
}

@article{li2025eagle,
  title={Eagle-3: Scaling up inference acceleration of large language models via training-time test},
  author={Li, Yuhui and Wei, Fangyun and Zhang, Chao and Zhang, Hongyang},
  journal={arXiv preprint arXiv:2503.01840},
  year={2025}
}

@article{dubey2024llama,
  title={The llama 3 herd of models},
  author={Dubey, Abhimanyu and Jauhri, Abhinav and Pandey, Abhinav and Kadian, Abhishek and Al-Dahle, Ahmad and Letman, Aiesha and Mathur, Akhil and Schelten, Alan and Yang, Amy and Fan, Angela and others},
  journal={arXiv e-prints},
  pages={arXiv--2407},
  year={2024}
}

@inproceedings{bai2024longbench,
  title={Longbench: A bilingual, multitask benchmark for long context understanding},
  author={Bai, Yushi and Lv, Xin and Zhang, Jiajie and Lyu, Hongchang and Tang, Jiankai and Huang, Zhidian and Du, Zhengxiao and Liu, Xiao and Zeng, Aohan and Hou, Lei and others},
  booktitle={Proceedings of the 62nd Annual Meeting of the Association for Computational Linguistics (Volume 1: Long Papers)},
  pages={3119--3137},
  year={2024}
}

@misc{huang2021efficient,
      title={Efficient Attentions for Long Document Summarization}, 
      author={Luyang Huang and Shuyang Cao and Nikolaus Parulian and Heng Ji and Lu Wang},
      year={2021},
      eprint={2104.02112},
      archivePrefix={arXiv},
      primaryClass={cs.CL}
    }

@article{cobbe2021gsm8k,
  title={Training Verifiers to Solve Math Word Problems},
  author={Cobbe, Karl and Kosaraju, Vineet and Bavarian, Mohammad and Chen, Mark and Jun, Heewoo and Kaiser, Lukasz and Plappert, Matthias and Tworek, Jerry and Hilton, Jacob and Nakano, Reiichiro and Hesse, Christopher and Schulman, John},
  journal={arXiv preprint arXiv:2110.14168},
  year={2021}
}

@inproceedings{
hendrycks2021measuring,
title={Measuring Mathematical Problem Solving With the {MATH} Dataset},
author={Dan Hendrycks and Collin Burns and Saurav Kadavath and Akul Arora and Steven Basart and Eric Tang and Dawn Song and Jacob Steinhardt},
booktitle={Thirty-fifth Conference on Neural Information Processing Systems Datasets and Benchmarks Track (Round 2)},
year={2021},
url={https://openreview.net/forum?id=7Bywt2mQsCe}
}

@misc{tang2024mathscalescalinginstructiontuning,
      title={MathScale: Scaling Instruction Tuning for Mathematical Reasoning}, 
      author={Zhengyang Tang and Xingxing Zhang and Benyou Wang and Furu Wei},
      year={2024},
      eprint={2403.02884},
      archivePrefix={arXiv},
      primaryClass={cs.CL},
      url={https://arxiv.org/abs/2403.02884}, 
}

@inproceedings{lu-etal-2024-experts,
    title = "Not All Experts are Equal: Efficient Expert Pruning and Skipping for Mixture-of-Experts Large Language Models",
    author = "Lu, Xudong  and
      Liu, Qi  and
      Xu, Yuhui  and
      Zhou, Aojun  and
      Huang, Siyuan  and
      Zhang, Bo  and
      Yan, Junchi  and
      Li, Hongsheng",
    editor = "Ku, Lun-Wei  and
      Martins, Andre  and
      Srikumar, Vivek",
    booktitle = "Proceedings of the 62nd Annual Meeting of the Association for Computational Linguistics (Volume 1: Long Papers)",
    month = aug,
    year = "2024",
    address = "Bangkok, Thailand",
    publisher = "Association for Computational Linguistics",
    url = "https://aclanthology.org/2024.acl-long.334/",
    doi = "10.18653/v1/2024.acl-long.334",
    pages = "6159--6172"
}

@article{guo2025deepseek,
  title={Deepseek-r1: Incentivizing reasoning capability in llms via reinforcement learning},
  author={Guo, Daya and Yang, Dejian and Zhang, Haowei and Song, Junxiao and Zhang, Ruoyu and Xu, Runxin and Zhu, Qihao and Ma, Shirong and Wang, Peiyi and Bi, Xiao and others},
  journal={arXiv preprint arXiv:2501.12948},
  year={2025}
}

@misc{shazeer2017outrageouslylargeneuralnetworks,
      title={Outrageously Large Neural Networks: The Sparsely-Gated Mixture-of-Experts Layer}, 
      author={Noam Shazeer and Azalia Mirhoseini and Krzysztof Maziarz and Andy Davis and Quoc Le and Geoffrey Hinton and Jeff Dean},
      year={2017},
      eprint={1701.06538},
      archivePrefix={arXiv},
      primaryClass={cs.LG},
      url={https://arxiv.org/abs/1701.06538}, 
}

@misc{fedus2022switchtransformersscalingtrillion,
      title={Switch Transformers: Scaling to Trillion Parameter Models with Simple and Efficient Sparsity}, 
      author={William Fedus and Barret Zoph and Noam Shazeer},
      year={2022},
      eprint={2101.03961},
      archivePrefix={arXiv},
      primaryClass={cs.LG},
      url={https://arxiv.org/abs/2101.03961}, 
}

@misc{rajbhandari2022deepspeedmoeadvancingmixtureofexpertsinference,
      title={DeepSpeed-MoE: Advancing Mixture-of-Experts Inference and Training to Power Next-Generation AI Scale}, 
      author={Samyam Rajbhandari and Conglong Li and Zhewei Yao and Minjia Zhang and Reza Yazdani Aminabadi and Ammar Ahmad Awan and Jeff Rasley and Yuxiong He},
      year={2022},
      eprint={2201.05596},
      archivePrefix={arXiv},
      primaryClass={cs.LG},
      url={https://arxiv.org/abs/2201.05596}, 
}

@misc{kwon2022fastposttrainingpruningframework,
      title={A Fast Post-Training Pruning Framework for Transformers}, 
      author={Woosuk Kwon and Sehoon Kim and Michael W. Mahoney and Joseph Hassoun and Kurt Keutzer and Amir Gholami},
      year={2022},
      eprint={2204.09656},
      archivePrefix={arXiv},
      primaryClass={cs.CL},
      url={https://arxiv.org/abs/2204.09656}, 
}
